\documentclass[runningheads]{llncs}
\usepackage[T1]{fontenc}
\usepackage{graphicx,verbatim}
\usepackage{hyperref}
\usepackage{amsmath}
\usepackage{amssymb}
\usepackage[table]{xcolor} 
\usepackage{booktabs} 
\usepackage{multirow} 
\usepackage{rotating} 
\usepackage{marvosym}
\definecolor{oran_tab}{RGB}{252, 242, 237}
\definecolor{blue_tab}{RGB}{227, 240, 251}
\definecolor{green_tab}{RGB}{240, 255, 240}

\begin{document}
\title{Probe-EM: Targeted Neuron Tracing via Training-Free Semantic 
Verification}
\author{
	Liuyun Jiang\inst{1,2}\textsuperscript{*} \and
	Yanchao Zhang\inst{1,2}\textsuperscript{*} \and
	Jinyue Guo\inst{1,3} \and
	Chuanyue Chen\inst{1,4} \and
	Haiyang Yan\inst{1,2} \and
	Ye Yuan\inst{1,4} \and
	Jing Liu\inst{1}\textsuperscript{(\Letter)} \and
	Hua Han\inst{1,2}\textsuperscript{(\Letter)}
}
\authorrunning{L. Jiang et al.}

\institute{
	State Key Laboratory of Brain Cognition and Brain-inspired Intelligence 
	Technology,
	Institute of Automation, Chinese Academy of Sciences, Beijing, China\\
	\email{liujing2016@ia.ac.cn, hua.han@ia.ac.cn}
	\and
	School of Future Technology, University of Chinese Academy of Sciences
	\and
	School of Artificial Intelligence, University of Chinese Academy of Sciences
	\and
	School of Advanced Interdisciplinary Sciences, University of Chinese 
	Academy of Sciences
}
\maketitle              
\begingroup
\renewcommand\thefootnote{*}
\footnotetext{These authors contributed equally.}
\endgroup
\begin{abstract}
Establishing large-scale, high-resolution neural connectivity maps is 
fundamental to elucidating the structural basis of brain function. However, 
when processing terabyte- or petabyte-scale electron microscopy data, 
over-segmentation inherent in automated reconstruction algorithms remains a 
critical bottleneck, requiring extensive manual proofreading spanning 
person-years. To alleviate the heavy reliance on annotated data and the 
limited flexibility of conventional tracing methods, we propose a 
training-free, targeted neuron tracing framework. Specifically, we introduce a 
skeleton-guided Heuristic Spatial Search paradigm that leverages geometric 
priors to iteratively reconstruct neuronal morphologies through a 
probing–verification cycle. To achieve robust zero-shot semantic verification, 
we further develop a Dimension-Aware Semantic Verification strategy built upon 
the foundation model NeuroSAM 2. This strategy resolves intra-slice 
splits via Planar Ensemble Consensus and inter-slice splits via Axial 
Spatio-Temporal Propagation. Notably, we integrate the proposed workflow into 
the Neuroglancer visualization platform, enabling an interactive 
human-in-the-loop proofreading system. Experimental results demonstrate that 
the proposed method outperforms supervised baselines and reduces manual 
proofreading time by 33.4\%. The source code is publicly available at 
\url{https://github.com/HeadLiuYun/Probe-EM}.

\end{abstract}
\section{Introduction}
Connectomics aims to elucidate the structural foundations of brain function by 
reconstructing high-fidelity neural connectivity maps at nanometer resolution 
\cite{a:6,a:5,a:8}. Driven by recent breakthroughs in electron microscopy (EM) 
imaging, connectomic datasets have rapidly expanded from localized circuits to 
whole-brain volumes at the terabyte-to-petabyte scale 
\cite{microns,flywire,h01}. 
Nevertheless, current automated reconstruction methods 
\cite{mala,embedded,ffn,superhuman,lsd} 
remain prone to over-segmentation errors, necessitating prohibitively 
labor-intensive 
manual proofreading to recover complete neuronal morphologies and connectivity. 
Given that proofreading a single \textit{Drosophila} brain requires tens of 
person-years of human effort \cite{tenyear1,tenyear2}, developing efficient 
automated 
validation methods has become imperative to address scalability bottlenecks in 
modern connectomics \cite{2025method}.

To alleviate the substantial manual effort required for proofreading, a range 
of automated validation strategies has been developed to address 
over-segmentation in neuronal reconstructions. Existing frameworks primarily 
employ deep learning models to infer connectivity probabilities between neurite 
segments. For instance, Matejek et al. \cite{edgecnn} utilized 3D CNNs on 
biologically constrained graphs to predict segment merges, while Chen et al. 
\cite{multimodal} leveraged multimodal contrastive learning to train 
high-performance classifiers. Similarly, RoboEM \cite{roboem} introduced a 
self-steering system simulating flight navigation to connect segments in 3D 
space. Despite these advances, existing methods are hindered by two critical 
limitations. \textbf{(1) Rigid validation paradigm.} Most existing frameworks 
model error correction as a binary inference task between discrete segments. 
This approach necessitates exhaustive global scanning, which lacks the 
necessary flexibility for targeted neurite tracing. In practical scenarios 
where researchers prioritize specific neuronal populations, existing methods 
cannot initiate tracing directly from user-defined seed segments to reconstruct 
complete morphologies, resulting in substantial computational resources and 
time being unnecessarily consumed on non-target regions. \textbf{(2) Heavy 
reliance on annotated data.} Existing models depend heavily on large-scale 
ground-truth labels for supervised training, limiting their out-of-the-box 
generalizability when applied to novel datasets with distinct characteristics.

We propose a training-free, targeted neuron tracing framework to address these 
challenges. Specifically, the framework implements a Heuristic Spatial Search 
(HSS) paradigm that emulates human expert decision-making logic. This paradigm 
leverages skeleton-based geometric priors to autonomously explore and validate 
fragments within the spatial vicinity of a seed segment, thereby enabling 
efficient local-to-global topological reconstruction. This strategy avoids the 
computational redundancy inherent in conventional global scanning approaches 
for non-target regions. To reduce the dependency on large-scale annotations 
required by existing supervised methods, we develop a Dimension-Aware Semantic 
Verification (DASV) module, anchored by \textbf{NeuroSAM 2}, a foundation model 
\cite{sam2} specifically fine-tuned for neuronal EM imagery. To ensure robust 
connectivity inference, the DASV module handles intra-slice and inter-slice 
split scenarios using a Planar Ensemble Consensus (PEC) and an Axial 
Spatio-Temporal Propagation (ASP) mechanism, respectively. Finally, we 
integrate the proposed algorithm into the Neuroglancer 
\footnote{https://github.com/google/neuroglancer} visualization platform to 
construct a human-in-the-loop interactive proofreading system.

The primary contributions of this work are summarized as follows: 
(1) We propose the HSS tracing paradigm, which leverages skeleton geometric 
priors to guide targeted neurite growth, thereby eliminating the computational 
redundancy inherent in conventional global scanning approaches. 
(2) We introduce NeuroSAM 2 and the DASV strategy to achieve training-free 
automated validation across both intra-slice and inter-slice split scenarios. 
(3) Experiments on the SCN dataset demonstrate that our framework outperforms 
supervised baselines, reducing manual proofreading time by 33.4\% while 
maintaining high reconstruction fidelity.
\begin{figure*}[t]
	\centering
	\includegraphics[width=1.0\textwidth]{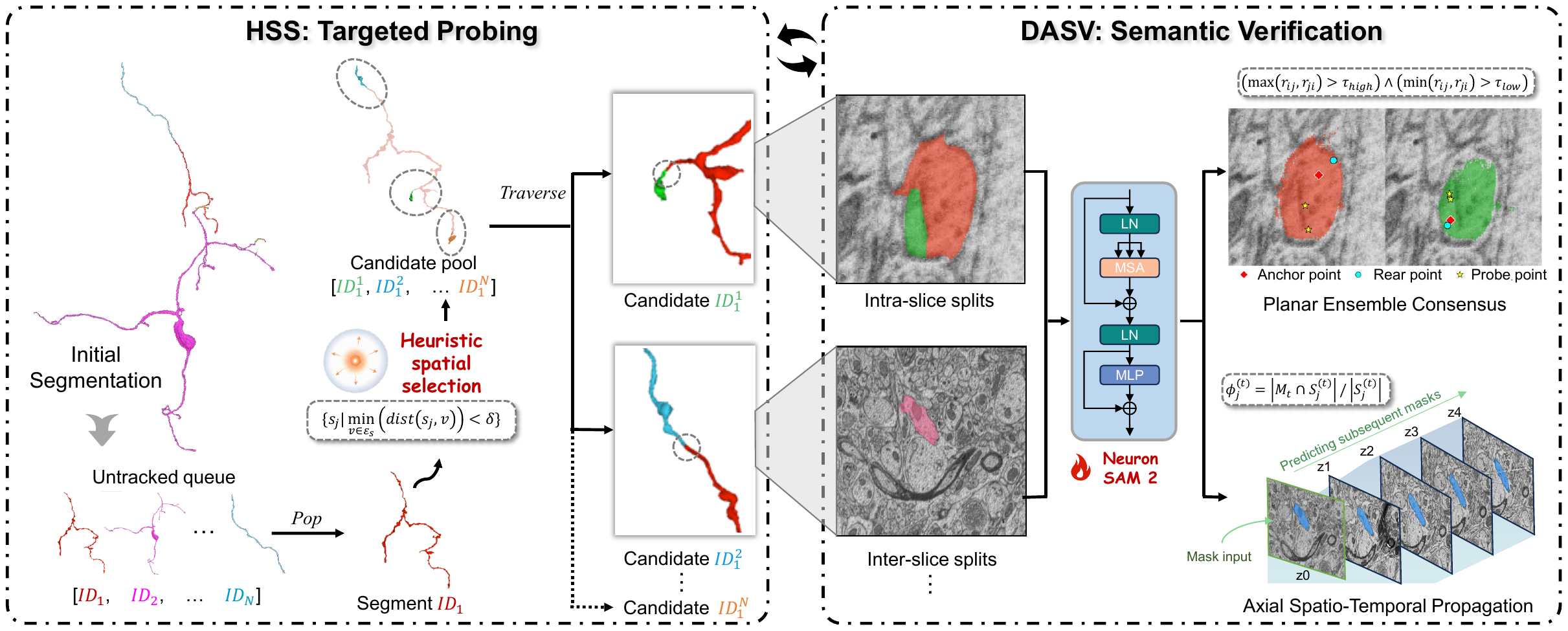} 
	\caption{Overview of Probe-EM. The framework facilitates iterative tracing 
	of complete neuronal morphologies through a synergistic 
	probing-verification cycle.}
	\label{fig1}
\end{figure*}

\section{Method}
As illustrated in Fig. \ref{fig1}, this work introduces a seed-guided active 
tracing paradigm that circumvents the computational redundancy inherent in 
conventional global approaches within non-target regions. The proposed 
framework incorporates an HSS module for targeted candidate probing alongside a 
DASV module for multi-dimensional connectivity inference. By leveraging the 
iterative refinement of a probing-verification cycle, our approach achieves 
efficient, high-fidelity reconstruction of neuronal topology.

\subsection{HSS: Targeted Probing}
We reformulate the neuron tracing problem as a targeted, seed-driven retrieval 
task. 
Let $\mathcal{V}$ denote the set of all segments within a volume, and let a 
complete 
biological neuron $N$ be represented by a subset of segments $\mathcal{S}_N 
\subseteq \mathcal{V}$, 
defined as $\mathcal{S}_N = \{ s^{(N)}_1, s^{(N)}_2, \dots, s^{(N)}_n \}$. 
Given a specific seed segment $s_{seed} \in \mathcal{S}_N$, the goal is to 
efficiently recover 
the remaining components $\mathcal{S}_N \setminus \{s_{seed}\}$ through 
iterative local exploration. 
Unlike conventional methods that rely on exhaustive pairwise connectivity 
predictions 
across the entire volume $(s_i, s_j) \in \mathcal{V} \times \mathcal{V}$, our 
targeted formulation 
naturally enables flexible, seed-driven tracing, prioritizing specific neuronal 
populations.

A processing queue $Q$ is maintained, initialized with a seed segment 
$s_{seed}$. In each iteration, the algorithm dequeues a segment $s$ and applies 
a skeletonization operator $\Gamma(s)$ to extract its spatial topology, 
represented as a graph $G_s = (V, E)$. Given that neuronal discontinuities 
predominantly occur at skeleton terminals, we explore adjacent fragments within 
spherical neighborhoods centered at each endpoint in the terminal set 
$\mathcal{E}_s = \{ v \in V \mid \text{deg}(v)=1 \}$. The candidate set 
$\mathcal{C}$ for the subsequent DASV module is formally defined as:
\begin{equation}
	\mathcal{C} = \{ s_j \in \mathcal{V} \mid \min_{v \in \mathcal{E}_s} \| 
	\text{pos}(s_j) - \text{pos}(v) \|_2 < \delta \}
\end{equation}
where $\text{pos}(\cdot)$ denotes 3D coordinates and $\delta = 500 \, 
\text{nm}$ is the search radius, ensuring sufficient coverage for potential 
continuities.

To mitigate error propagation and prevent computational divergence, a 
topological pruning mechanism is incorporated into the HSS module. During the 
tracing cycle, the algorithm continuously monitors the topological complexity 
of the growth front. When a newly detected segment exhibits an excessive number 
of endpoints or abnormal skeleton endpoint density---morphological signatures 
typically associated with glial cell misconnections or reconstruction 
mergers---it is identified as a topological anomaly, and the expansion of that 
branch is preemptively terminated. Segments that pass this topological 
validation are subsequently integrated into the global topology and appended to 
the queue $Q$ for further exploration. This procedure iterates until $Q$ is 
empty, thereby achieving a comprehensive reconstruction of the neuronal 
topological morphology.

\subsection{DASV: Semantic Verification}
The DASV module evaluates the candidate set $\mathcal{C}$ generated by HSS, 
leveraging multi-dimensional semantic features to verify inter-segment 
connectivity. Candidate pairs are categorized into two distinct scenarios: 
\textbf{intra-slice splits} occurring within the same XY plane, and 
\textbf{inter-slice splits} spanning multiple Z-axis slices. This 
dimension-aware classification enables specialized verification strategies 
tailored to the specific reconstruction challenges inherent in each scenario.

\noindent\textbf{Planar Ensemble Consensus.}
In intra-slice scenarios, the principal challenge lies in discerning genuine 
semantic boundaries between candidate segments. To address this, we leverage 
NeuroSAM 2, a foundation model fine-tuned on neuronal EM datasets, to perform 
connectivity verification via a bi-directional ensemble consensus task.

Specifically, the procedure begins with a robust geometric prompt sampling 
strategy. For a candidate pair $(s_i, s_j)$, Euclidean distance transforms are 
utilized to extract three types of complementary prompt points shown in Fig. 
\ref{fig1}: (1) anchor points at morphological centers to establish segment 
identity; (2) probe points positioned in close proximity to the target segment 
to test boundary permeability; and (3) rear points located distal to the target 
to ensure that NeuroSAM 2 maintains the morphological integrity of the segment 
during mask expansion. During the verification stage, the algorithm executes 
$K=5$ independent randomized trials. Each trial performs a cross-prediction: 
predicting $s_j$ given prompts from $s_i$, and vice versa, to calculate the 
bidirectional overlap ratios $r_{ij}$ and $r_{ji}$. A trial is considered 
successful only if the scores satisfy a dual-threshold criterion:
\begin{equation}
	(\max(r_{ij}, r_{ji}) > \tau_{\text{high}}) \land (\min(r_{ij}, r_{ji}) > 
	\tau_{\text{low}})
\end{equation}
where $\tau_{\text{high}}$ and $\tau_{\text{low}}$ enforce both strong merging 
evidence and minimal bidirectional agreement. Final connectivity is confirmed 
only when an ensemble consensus is achieved (e.g., passing 4 out of 5 trials). 
This stringent voting logic leverages the boundary awareness of NeuroSAM 2 
while effectively suppressing false connections resulting from mask leakage or 
sampling stochasticity.

\noindent\textbf{Axial Spatio-Temporal Propagation.}
Verifying inter-slice axial splits poses significant challenges, as these 
discontinuities frequently span multi-slice Z-axis gaps. Conventional methods 
typically rely on Intersection over Union (IoU) matching between adjacent slice 
masks. However, such heuristics are highly susceptible to slice misalignments, 
tissue loss, or drastic morphological shifts of neurites, thereby failing to 
maintain long-range topological robustness. To address these limitations, we 
leverage the long-range modeling capabilities of NeuroSAM 2 to introduce an 
axial spatio-temporal propagation mechanism.

Specifically, the algorithm formulates the 3D neighborhood of a split as a 
pseudo-video sequence aligned along the Z-axis. Using the mask of the source 
segment as the initial prompt, the convolutional memory module within NeuroSAM 
2 is driven to autonomously propagate and evolve the segment identity along the 
axial direction for $N$ frames. In contrast to static IoU matching, this 
memory-augmented propagation effectively captures and adapts to the 
morphological evolution of neurites across consecutive slices. During the 
$N$-frame mask evolution powered by NeuroSAM 2, the algorithm dynamically 
monitors the spatial overlap between the prediction and potential candidates 
through volumetric occupancy detection. 

Formally, let $M_t$ and $S_j^{(t)}$ denote the predicted mask and the voxel set 
of candidate segment $s_j$ at frame $t$, respectively. The instantaneous 
connectivity score is computed as $\phi_j^{(t)} = |M_t \cap S_j^{(t)}| / 
|S_j^{(t)}|$. To identify the most robust connection throughout the traversal, 
the maximum coverage ratio across the propagation sequence is extracted as the 
final score: $\Phi_j = \max_{t \in \{1,\dots,N\}} \phi_j^{(t)}$. Long-range 
topological continuity is confirmed if $\Phi_j$ exceeds a predefined semantic 
collision threshold $\tau_{\text{occ}}$.

Finally, the HSS and DASV modules are integrated into the Neuroglancer 
visualization platform, creating an interactive, human-in-the-loop 
proofreading system.

\section{Experiments}
\subsection{Datasets and Evaluation Metrics}
We utilized the first complete serial section electron microscopy (ssEM) 
dataset of a unilateral mouse suprachiasmatic nucleus (SCN) \cite{SCN}. With a 
voxel resolution of $5 \times 5 \times 40 \text{ nm}^3$, the dataset comprises 
6,824 
serial sections and covers a physical volume of approximately $0.06 \text{ 
mm}^3$, amounting to a raw data size of 88.3~TB. The SCN dataset was processed 
using a 3D Residual U-Net \cite{superhuman} to generate affinity maps, followed 
by the distance transform watershed algorithm \cite{DTW} to obtain the initial 
segmentation results. The SCN dataset is characterized by prominent cable-like 
axon fascicles traversing the nucleus. However, these slender and elongated 
structures are inherently susceptible to over-segmentation errors during 
automated 3D reconstruction. We annotated a total of 36 axon fascicles and 12 
soma-containing neurons. Of these, 12 axon fascicles and 4 soma-containing 
neurons were selected to train the learning-based baseline methods 
\cite{multimodal,edgecnn}, while the 
remainder constituted the test set.

To quantitatively evaluate the tracing framework, we adopt a 
skeleton-node-based evaluation metric. Specifically, Recall, Precision, and F1 
scores are computed by comparing the skeleton nodes covered by the retrieved 
segments against the ground-truth nodes. Crucially, nodes belonging to the 
initial seed segment are excluded from all metric calculations to strictly 
assess the algorithm's incremental tracing efficacy.

\subsection{Implementation Details}
Several learning-based baseline methods \cite{multimodal,edgecnn,pointnet} were 
implemented for comparative evaluation. During training, all baseline models 
received local $120 \times 120 \times 30$ volumetric patches centered at 
potential junctions as input. These models were optimized via binary 
cross-entropy using the Adam optimizer (learning rate $1 \times 10^{-4}$) for 
50 epochs. A WeightedRandomSampler was employed to mitigate the class imbalance 
(433 positive versus 8,699 negative pairs). Unlike baselines requiring 
supervised connectivity training, our tracing 
framework is training-free. NeuroSAM 2 was fine-tuned solely on the EMNeuron 
dataset \cite{segneuron} without connectivity-specific supervision. For DASV 
tracing, we empirically set $\tau_{\text{high}}=0.5$, $\tau_{\text{low}}=0.1$, 
$\tau_{\text{occ}}=0.6$, and the axial propagation window size $N=5$. 
Experiments were conducted on an NVIDIA RTX 4090 GPU.

\subsection{Results}
\begin{figure*}[t]
	\centering
	\includegraphics[width=1\textwidth]{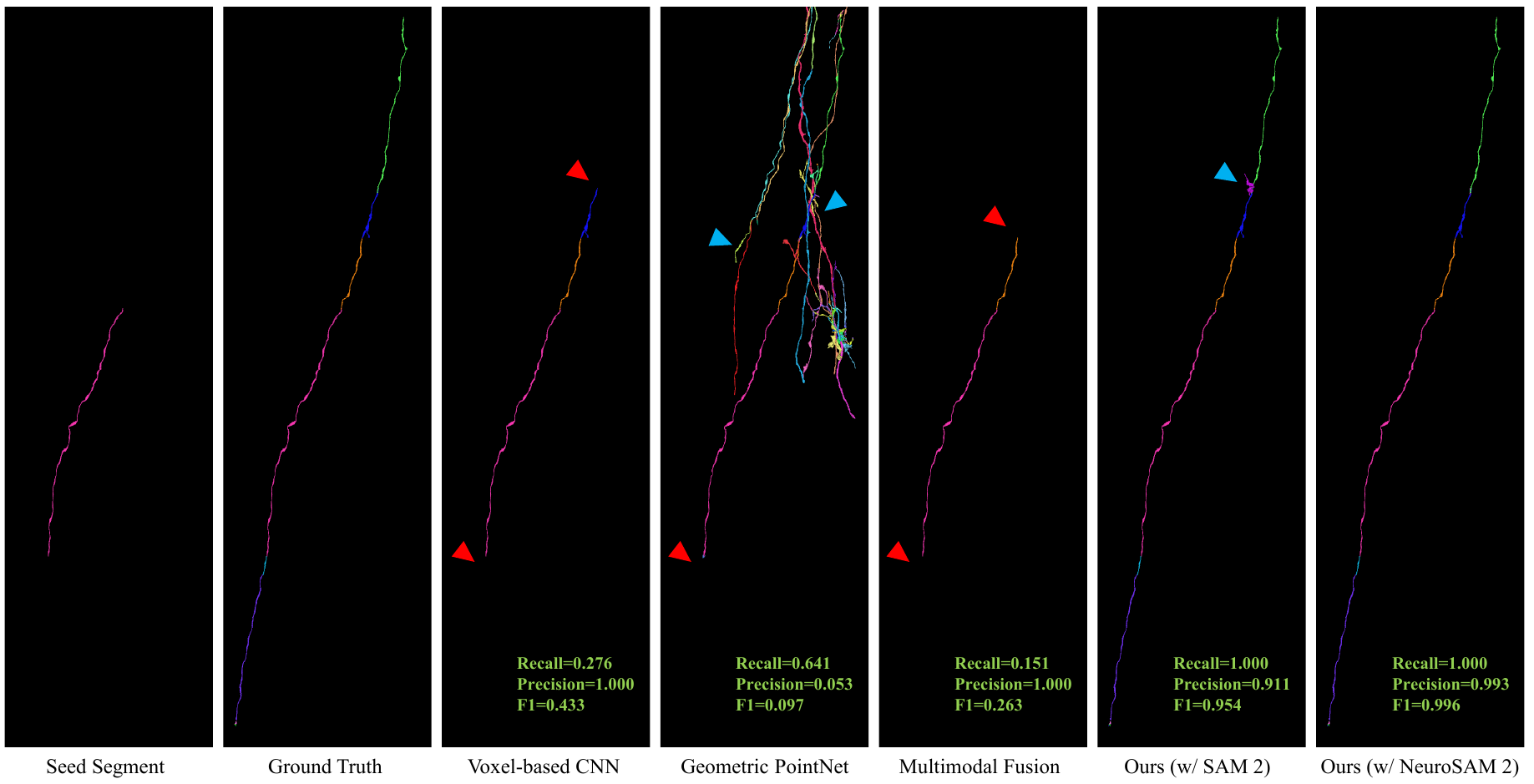} 
	\caption{Qualitative visualization of neuron tracing results. The leftmost 
	Seed Segment serves as the input for comparison against the Ground Truth 
	and various methods. Red and blue triangles highlight regions of incomplete 
	tracing (missed neurites) and erroneous tracking (spurious fragments), 
	respectively.}
	\label{fig2}
\end{figure*}

\noindent\textbf{Neuron tracking performance.}
Experiments are categorized into two distinct tasks: \textbf{Axon Fascicle 
Tracing}, which aims to reconstruct the complete morphology starting from a 
localized seed segment within an axon bundle, and \textbf{Soma-seeded Tracing}, 
which focuses on the reconstruction of full neuronal structures originating 
from a soma-containing segment. For comparative evaluation, we benchmark our 
method against connectivity discrimination approaches based on voxel-wise 
morphological learning \cite{edgecnn}, point-cloud geometric learning 
\cite{pointnet}, and a multimodal framework \cite{multimodal} that integrates 
both representations. Additionally, we evaluate the impact of the foundation 
model by contrasting the off-the-shelf SAM 2 \cite{sam2} with our fine-tuned 
NeuroSAM 2 within the proposed framework. The quantitative results, summarized 
in Table \ref{tab1}, demonstrate that our method achieves the best overall 
performance, significantly outperforming the supervised learning-based 
baselines. We attribute this superiority to the inherent limitation of 
supervised approaches: they typically require extensive annotated pairs to 
capture the drastic morphological variations of segment cross-sections. In 
contrast, our approach leverages the robust generalization capabilities of the 
foundation model to handle such diversity effectively without heavy reliance on 
task-specific training data. These results are visually demonstrated in Fig. 
\ref{fig2}, 
where our method achieves superior topological integrity.
\begin{table*}[t]
	\centering
	\small
	\renewcommand{\arraystretch}{0.8}
	\setlength{\tabcolsep}{6.0pt}
	\caption{Quantitative performance comparison of neuron tracing on the SCN 
	dataset.}
	\label{tab1}
	\begin{tabular}{l|ccc|ccc}
		\toprule
		\multirow{2}{*}{\textbf{Methods}} & \multicolumn{3}{c|}{\textbf{Axon 
		Fascicle Tracing}} & \multicolumn{3}{c}{\textbf{Soma-seeded Tracing}} 
		\\ 
		\cmidrule{2-7} 
		& Recall & Precision & F1 & Recall & Precision & F1 \\ 
		\midrule        
		\multicolumn{7}{l}{\textit{Learning-based Methods}} \\
		Voxel-based CNN \cite{edgecnn} & 0.542 & \textbf{0.902} & 0.571 & 0.245 
		& 0.428 
		& 0.293 
		\\ 
		Geometric PointNet \cite{pointnet} & 0.509 & 0.485 & 0.288 & 0.181 & 
		0.498 & 
		0.210 \\ 
		Multimodal Fusion \cite{multimodal} & 0.421 & 0.650 & 0.436 & 0.160 & 
		0.372 & 
		0.202 \\
		\midrule
		\multicolumn{7}{l}{\textit{Training-free Methods}} \\
		Ours (w/ SAM 2) & 0.736 & 0.494 & 0.467 & 0.492 & 0.312 & 0.310 \\
		Ours (w/ NeuroSAM 2) & \textbf{0.753} & 0.626 & \textbf{0.595} & 
		\textbf{0.705} & \textbf{0.565} & \textbf{0.544} \\
			
		\bottomrule
	\end{tabular}
\end{table*}
\begin{table}[t]
	\centering
	\small
	\renewcommand{\arraystretch}{0.8}
	\setlength{\tabcolsep}{5pt}
	\caption{Quantitative comparison of annotation efficiency and accuracy 
		between manual-only and Probe-EM-assisted workflows.}
	\label{ta2}
	\begin{tabular}{c|cc|cc|cc}
		\toprule
		\multirow{2}{*}{\textbf{Annotator}} & 
		\multicolumn{2}{c|}{\textbf{Manual Only}} & 
		\multicolumn{2}{c|}{\textbf{w/ Probe-EM}} & 
		\multicolumn{2}{c}{\textbf{Improvement}} \\ 
		\cmidrule{2-7}
		& Time (min) & F1 Score & Time (min) & F1 Score & $\Delta$ Time & 
		$\Delta$ F1 \\ 
		\midrule
		Annotator 1 & 52.6 & 0.874 & 29.2 & 0.892 & -44.5\% & +2.1\% \\
		Annotator 2 & 31.6 & 0.827 & 20.0 & 0.974 & -36.7\% & +17.8\% \\
		Annotator 3 & 30.2 & 0.894 & 24.5 & 0.899 & -18.9\% & +0.6\% \\
		\midrule
		\textbf{Average} & \textbf{38.1} & \textbf{0.865} & \textbf{24.6} & 
		\textbf{0.922} & \textbf{-33.4\%} & \textbf{+6.8\%} \\
		\bottomrule
	\end{tabular}
\end{table}

\noindent\textbf{Human-in-the-loop Efficiency Study.}
To assess the practical impact of our framework on the proofreading workflow, 
we conducted a comparative user study involving three experienced annotators, 
each assigned to proofread five distinct neurons. As detailed in Table 
\ref{ta2}, the integration of Probe-EM significantly expedited the neurite 
proofreading process compared to the manual-only baseline. On average, the 
proofreading time was reduced from 38.1 to 24.6 minutes, yielding a 33.4\% 
efficiency gain. Crucially, this acceleration was accompanied by a concurrent 
improvement in tracing quality, with the mean F1 score increasing by 6.8\%. 
These results indicate that the suggestions provided by Probe-EM not only 
mitigate the manual annotation burden but also effectively assist annotators in 
identifying branches that are frequently overlooked during manual 
slice-by-slice inspection.
\begin{table}[h]
	\centering
	\small
	\renewcommand{\arraystretch}{0.8}
	\setlength{\tabcolsep}{8pt} 
	\caption{Ablation study of core components and hyperparameters. GPS denotes 
		Geometric Prompt Sampling, and $N$ represents the axial propagation 
		window 
		size.}
	\label{tab3}
	\begin{tabular}{l|cccc|ccc} 
		\toprule
		Configuration & PEC & ASP & GPS & $N$ & Recall & Precision & F1 \\ 
		\midrule
		w/o PEC &  & \checkmark & \checkmark & 5 & 0.595 & 0.605 & 0.483 \\
		w/o ASP & \checkmark &  & \checkmark & 5 & 0.020 & \textbf{0.800} & 
		0.036 \\
		w/o GPS & \checkmark & \checkmark &  & 5 & 0.588 & 0.346 & 0.328 \\    
		\midrule
		Ours ($N=3$) & \checkmark & \checkmark & \checkmark & 3 & 0.611 & 0.624 
		& 0.484 \\
		Ours (Default) & \checkmark & \checkmark & \checkmark & 5 & 
		\textbf{0.691} & 0.639 & \textbf{0.586} \\
		Ours ($N=8$) & \checkmark & \checkmark & \checkmark & 8 & 0.589 & 0.396 
		& 0.371 \\
		\bottomrule
	\end{tabular}
\end{table}

\noindent\textbf{Ablation Studies.}
To evaluate the efficacy of the core components within our framework, 
comprehensive ablation experiments were conducted on five distinct neurons from 
each task, as summarized in Table \ref{tab3}. These results demonstrate that 
both the ASP and PEC modules are indispensable. Notably, removing the ASP 
module leads to a precipitous drop in recall, indicating that most neurite 
discontinuities occur along the z-axis and underscoring the necessity of 
inter-slice propagation. Furthermore, the Geometric Prompt Sampling strategy 
significantly outperforms naive center-point prompting, as the strategic 
incorporation of probe and rear points better preserves the semantic integrity 
of the segments. Finally, we evaluate the impact of the axial propagation 
window size $N$. A smaller window limits the capacity to bridge extensive 
spatial gaps, whereas an excessively large window accumulates propagation 
errors and introduces semantic drift, ultimately degrading precision.

\section{Conclusion}
We proposed Probe-EM, a training-free tracing framework. By combining HSS with 
NeuroSAM 2-based DASV, our method improves performance while reducing manual 
proofreading time by 33.4\%, providing efficient support for complex neuron 
reconstruction and validation workflows in large-scale connectomics.

\bibliographystyle{splncs04}
\bibliography{reference}

@article{microns,
	title={Functional connectomics spanning multiple areas of mouse visual 
	cortex},
	journal={Nature},
	volume={640},
	number={8058},
	pages={435--447},
	year={2025},
	publisher={Nature Publishing Group UK London}
}

@article{flywire,
	title={Whole-brain annotation and multi-connectome cell typing of 
	Drosophila},
	author={Schlegel, Philipp and Yin, Yijie and Bates, Alexander S and 
	Dorkenwald, Sven and Eichler, Katharina and Brooks, Paul and Han, Daniel S 
	and Gkantia, Marina and Dos Santos, Marcia and Munnelly, Eva J and others},
	journal={Nature},
	volume={634},
	number={8032},
	pages={139--152},
	year={2024},
	publisher={Nature Publishing Group UK London}
}

@article{h01,
	title={A petavoxel fragment of human cerebral cortex reconstructed at 
	nanoscale resolution},
	author={Shapson-Coe, Alexander and Januszewski, Micha{\l} and Berger, 
	Daniel R and Pope, Art and Wu, Yuelong and Blakely, Tim and Schalek, 
	Richard L and Li, Peter H and Wang, Shuohong and Maitin-Shepard, Jeremy and 
	others},
	journal={Science},
	volume={384},
	number={6696},
	pages={eadk4858},
	year={2024},
	publisher={American Association for the Advancement of Science}
}

@article{tenyear1,
	title={Neuronal wiring diagram of an adult brain},
	author={Dorkenwald, Sven and Matsliah, Arie and Sterling, Amy R and 
	Schlegel, Philipp and Yu, Szi-Chieh and McKellar, Claire E and Lin, Albert 
	and Costa, Marta and Eichler, Katharina and Yin, Yijie and others},
	journal={Nature},
	volume={634},
	number={8032},
	pages={124--138},
	year={2024},
	publisher={Nature Publishing Group UK London}
}

@article{tenyear2,
	title={A connectome and analysis of the adult Drosophila central brain},
	author={Scheffer, Louis K and Xu, C Shan and Januszewski, Michal and Lu, 
	Zhiyuan and Takemura, Shin-ya and Hayworth, Kenneth J and Huang, Gary B and 
	Shinomiya, Kazunori and Maitlin-Shepard, Jeremy and Berg, Stuart and 
	others},
	journal={elife},
	volume={9},
	pages={e57443},
	year={2020},
	publisher={eLife Sciences Publications, Ltd}
}

@article{superhuman,
	title={Superhuman accuracy on the SNEMI3D connectomics challenge},
	author={Lee, Kisuk and Zung, Jonathan and Li, Peter and Jain, Viren and 
	Seung, H Sebastian},
	journal={arXiv preprint arXiv:1706.00120},
	year={2017}
}

@article{mala,
	title={Large scale image segmentation with structured loss based deep 
	learning for connectome reconstruction},
	author={Funke, Jan and Tschopp, Fabian and Grisaitis, William and Sheridan, 
	Arlo and Singh, Chandan and Saalfeld, Stephan and Turaga, Srinivas C},
	journal={IEEE transactions on pattern analysis and machine intelligence},
	volume={41},
	number={7},
	pages={1669--1680},
	year={2018},
	publisher={IEEE}
}

@article{ffn,
	title={High-precision automated reconstruction of neurons with 
	flood-filling networks},
	author={Januszewski, Micha{\l} and Kornfeld, J{\"o}rgen and Li, Peter H and 
	Pope, Art and Blakely, Tim and Lindsey, Larry and Maitin-Shepard, Jeremy 
	and Tyka, Mike and Denk, Winfried and Jain, Viren},
	journal={Nature methods},
	volume={15},
	number={8},
	pages={605--610},
	year={2018},
	publisher={Nature Publishing Group US New York}
}

@inproceedings{edgecnn,
	title={Biologically-constrained graphs for global connectomics 
	reconstruction},
	author={Matejek, Brian and Haehn, Daniel and Zhu, Haidong and Wei, Donglai 
	and Parag, Toufiq and Pfister, Hanspeter},
	booktitle={Proceedings of the IEEE/CVF conference on computer vision and 
	pattern recognition},
	pages={2089--2098},
	year={2019}
}

@inproceedings{multimodal,
	title={Learning multimodal volumetric features for large-scale neuron 
	tracing},
	author={Chen, Qihua and Chen, Xuejin and Wang, Chenxuan and Liu, Yixiong 
	and Xiong, Zhiwei and Wu, Feng},
	booktitle={Proceedings of the AAAI Conference on Artificial Intelligence},
	volume={38},
	number={2},
	pages={1174--1182},
	year={2024}
}

@article{roboem,
	title={RoboEM: automated 3D flight tracing for synaptic-resolution 
	connectomics},
	author={Schmidt, Martin and Motta, Alessandro and Sievers, Meike and 
	Helmstaedter, Moritz},
	journal={Nature methods},
	volume={21},
	number={5},
	pages={908--913},
	year={2024},
	publisher={Nature Publishing Group US New York}
}

@article{SCN,
	title={Connectomic Organization of the Suprachiasmatic Nucleus},
	author={Liu, Jing and Yu, Jing and Shen, Lijun and Chen, Xi and Ma, Lei and 
	Zhai, Hao and Li, Linlin and Zhang, Lina and Wang, Lu and Yuan, Jingbin and 
	others},
	journal={bioRxiv},
	pages={2024--10},
	year={2024},
	publisher={Cold Spring Harbor Laboratory}
}

@article{DTW,
	title={Structure and function of a neocortical synapse},
	author={Holler, Simone and K{\"o}stinger, German and Martin, Kevan AC and 
	Schuhknecht, Gregor FP and Stratford, Ken J},
	journal={Nature},
	volume={591},
	number={7848},
	pages={111--116},
	year={2021},
	publisher={Nature Publishing Group UK London}
}

@inproceedings{segneuron,
	title={Segneuron: 3d neuron instance segmentation in any em volume with a 
	generalist model},
	author={Zhang, Yanchao and Guo, Jinyue and Zhai, Hao and Liu, Jing and Han, 
	Hua},
	booktitle={International Conference on Medical Image Computing and 
	Computer-Assisted Intervention},
	pages={589--600},
	year={2024},
	organization={Springer}
}

@inproceedings{pointnet,
	title={Pointnet: Deep learning on point sets for 3d classification and 
	segmentation},
	author={Qi, Charles R and Su, Hao and Mo, Kaichun and Guibas, Leonidas J},
	booktitle={Proceedings of the IEEE conference on computer vision and 
	pattern recognition},
	pages={652--660},
	year={2017}
}

@article{sam2,
	title={SAM 2: Segment Anything in Images and Videos},
	author={Ravi, Nikhila and Gabeur, Valentin and Hu, Yuan-Ting and Hu, 
	Ronghang and Ryali, Chaitanya and Ma, Tengyu and Khedr, Haitham and 
	R{\"a}dle, Roman and Rolland, Chloe and Gustafson, Laura and Mintun, Eric 
	and Pan, Junting and Alwala, Kalyan Vasudev and Carion, Nicolas and Wu, 
	Chao-Yuan and Girshick, Ross and Doll{\'a}r, Piotr and Feichtenhofer, 
	Christoph},
	journal={arXiv preprint arXiv:2408.00714},
	url={https://arxiv.org/abs/2408.00714},
	year={2024}
}

@article{lsd,
	title={Local shape descriptors for neuron segmentation},
	author={Sheridan, Arlo and Nguyen, Tri M and Deb, Diptodip and Lee, 
	Wei-Chung Allen and Saalfeld, Stephan and Turaga, Srinivas C and Manor, Uri 
	and Funke, Jan},
	journal={Nature methods},
	volume={20},
	number={2},
	pages={295--303},
	year={2023},
	publisher={Nature Publishing Group US New York}
}

@inproceedings{embedded,
	title={Learning to model pixel-embedded affinity for homogeneous instance 
	segmentation},
	author={Huang, Wei and Deng, Shiyu and Chen, Chang and Fu, Xueyang and 
	Xiong, Zhiwei},
	booktitle={Proceedings of the AAAI Conference on Artificial Intelligence},
	volume={36},
	number={1},
	pages={1007--1015},
	year={2022}
}

@article{a:5,
	title={BRAIN 2.0: Transforming neuroscience},
	author={Ngai, John},
	journal={Cell},
	volume={185},
	number={1},
	pages={4--8},
	year={2022},
	publisher={Elsevier}
}

@article{a:6,
	title={Whole-brain serial-section electron microscopy in larval zebrafish},
	author={Hildebrand, David Grant Colburn and Cicconet, Marcelo and Torres, 
	Russel Miguel and Choi, Woohyuk and Quan, Tran Minh and Moon, Jungmin and 
	Wetzel, Arthur Willis and Scott Champion, Andrew and Graham, Brett Jesse 
	and Randlett, Owen and others},
	journal={Nature},
	volume={545},
	number={7654},
	pages={345--349},
	year={2017},
	publisher={Nature Publishing Group UK London}
}

@article{a:8,
	title={Reconstruction of neocortex: Organelles, compartments, cells, 
	circuits, and activity},
	author={Turner, Nicholas L and Macrina, Thomas and Bae, J Alexander and 
	Yang, Runzhe and Wilson, Alyssa M and Schneider-Mizell, Casey and Lee, 
	Kisuk and Lu, Ran and Wu, Jingpeng and Bodor, Agnes L and others},
	journal={Cell},
	volume={185},
	number={6},
	pages={1082--1100},
	year={2022},
	publisher={Elsevier}
}

@article{2025method,
	title={Method of the Year 2025: electron microscopy-based connectomics},
	author={their Comment, In and Khajeh, Ramin},
	journal={nature methods},
	volume={22},
	pages={2463--2464},
	year={2025}
}

\end{document}